# Localization of a Unicycle-like Mobile Robot Using LRF and Omni-directional Camera

Tran Hiep Dinh, Manh Duong Phung, Thuan Hoang Tran, Quang Vinh Tran
VNU University of Engineering and Technology
Hanoi, Vietnam
tranhiep.dinh@vnu.edu.vn

*Abstract*— **This paper addresses the localization problem. The extended Kalman filter (EKF) is employed to localize a unicycle-like mobile robot equipped with a laser range finder (LRF) sensor and an omni-directional camera. The LRF is used to scan the environment which is described with line segments. The segments are extracted by a modified least square quadratic method in which a dynamic threshold is injected. The camera is employed to determine the robot's orientation. The prediction step of the EKF is performed by extracting parameters from the kinematic model and input signal of the robot. The correction step is conducted with the implementation of a line matching algorithm and the comparison between line's parameters of the local and global maps. In the line matching algorithm, a conversion matrix is introduced to reduce the computation cost. Experiments have been carried out in a real mobile robot system and the results prove the applicability of the method for the purpose of localization.**

## I. Introduction

The mapping and localization are fundamental to the automation of the mobile robot. The more accurately the robot could determine its position, the more autonomously it could operate to complete given tasks. In general, the robot determines its position based on feedback measurements from a sensor system such as the quadrature encoder, ultrasonic sensor, laser range finder (LRF) and visual camera. Nevertheless, the estimation based on data from one sensor is often not reliable enough to solve the localization and mapping problems. The multi-sensor data fusion therefore has received much attention recently due to its ability to combine advantages of each sensor type to extract more accurate information. Several combinations have been proposed such as the LRF and sonar, LRF and odometry, and LRF and CCD camera [1-3]. In [1], A. Diosi succeeded in combining a LRF with an advanced sonar sensor for better range and bearing measurements as well as classification of environment features. In [3], the operating environment was represented as a set of uncertain geometric features including segments, corners and semi planes based on data from the LRF and vision camera. These features were then fused together to extract the position and orientation of the mobile robot. A fusion approach to build a 2D-local map of the environment using the LRF and omni-directional camera is proposed in [4]. The core of the method is the development of a point to point matching algorithm. Though this approach is efficient and requires low computational cost, its accuracy may be vastly reduced if extracted feature points are not exactly maintained through detecting steps.

In this paper, the LRF and omni-directional camera are fused in a robot platform for the problem of localization. Instead of using features points as in [4], the LRF is employed to retrieve the structure of the environment, present it in form of line segments, match the segments with a global map and finally estimate the robot position. The camera is employed to determine the robot's orientation. Those data are then combined in an extended Kalman filter (EKF) to extract the *best* estimate of the robot's pose in a statistical sense. The main contribution of the paper is the proposal of a novel line matching algorithm with the present of a conversion matrix and a dynamic line-detection threshold. Simulations in MATLAB were carried out and experiments in a real robot system were implemented. The results confirm the effectiveness of the proposed fusion algorithm.

The paper is structured as follows. The localization algorithm is presented in section II. Section III describes the identification of Kalman filter's parameters from the LRF and omni-directional camera. Experiments are introduced in section IV. The paper concludes with an evaluation of the system.

## II. Localization Algorithm

In this work, the extended Kalman filter (EKF) is used as the localization algorithm. The EKF which is a set of recursive equations to estimate the state of a process in a way that minimizes the mean of the squared error has been proven to be effective for the robot localization [2, 8, 9, 10]. This section describes the implementation of the EKF to the mobile robot.

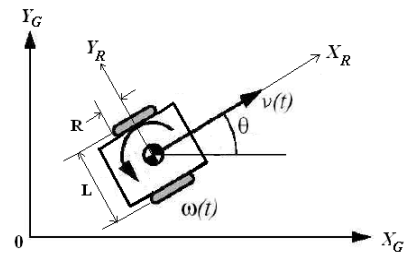

Figure 1. The robot's pose and parameters

The unicycle-like mobile robot with non-slipping and pure rolling is considered. Fig.1 shows the coordinate systems and notations for the robot, where $(X_G, Y_G)$ is the global coordinate, $(X_R, Y_R)$ is the local coordinate relative to the robot chassis. $R$ denotes the radius of driven wheels, and $L$ denotes the distance between the wheels.

Let $T_s$ be the sampling period, $\omega_L(k)$ and $\omega_R(k)$ be the measurements of rotational speeds of the left and right wheels with the encoders at the time $k$, respectively. The discrete kinematics model of the robot is given by:

$$x_{k+1} = x_k + \frac{R}{2}T_s(\omega_L(k)+\omega_R(k))\cos\theta_k$$
$$y_{k+1} = y_k + \frac{R}{2}T_s(\omega_L(k)+\omega_R(k))\sin\theta_k \quad (1)$$
$$\theta_{k+1} = \theta_k + \frac{R}{L}T_s(\omega_L(k)-\omega_R(k))$$

In the present of feedback measurements and disturbances, the system model is rewritten in state space representation as follows.

Let $\mathbf{x} = [x\ y\ \theta]^T$ be the state vector. This state can be observed by an absolute measurement, $\mathbf{z}$. This measurement is described by a nonlinear function, $h$, of the robot coordinates and an independent Gaussian noise process, $\mathbf{v}$. Denoting the function (1) as $f$, with an input vector $\mathbf{u}$ and a disturbance $\mathbf{w}$, the robot can be described as:

$$\mathbf{x}_{k+1} = f(\mathbf{x}_k, \mathbf{u}_k, \mathbf{w}_k)$$
$$\mathbf{z}_k = h(\mathbf{x}_k, \mathbf{v}_k) \quad (2)$$

where the random variables $\mathbf{w}_k$ and $\mathbf{v}_k$ represent the process and measurement noise respectively. They are assumed to be independent to each other, white, and with normal probability distributions: $\mathbf{w}_k \sim \mathbf{N}(0, \mathbf{Q}_k)$; $\mathbf{v}_k \sim \mathbf{N}(0, \mathbf{R}_k)$; $E(\mathbf{w}_i \mathbf{v}_j^T) = 0$. The steps to calculate the EKF are then performed through two phases: the prediction and correction as follows.

*A. The prediction step with time update equations:*

The prediction step updates the robot state based on the system model and input signal before the measurement is taken. Equations for this step are as below:

$$\hat{\mathbf{x}}_k^- = f(\hat{\mathbf{x}}_{k-1}, \mathbf{u}_{k-1}, \mathbf{0}) \quad (3)$$

$$\mathbf{P}_k^- = \mathbf{A}_k \mathbf{P}_{k-1} \mathbf{A}_k^T + \mathbf{W}_k \mathbf{Q}_{k-1} \mathbf{W}_k^T \quad (4)$$

where $\hat{\mathbf{x}}_k^- \in \mathfrak{R}^n$ is the *priori* state estimate at step $k$ given knowledge of the process prior to step $k$, $\hat{\mathbf{P}}_k^-$ denotes the covariance matrix of the state-prediction error, $\mathbf{A}_k$ is the Jacobian matrix of partial derivates of $f$ to $\mathbf{x}$:

$$\mathbf{A}_k = \frac{\partial f_k}{\partial \mathbf{x}}\bigg|_{(\hat{\mathbf{x}}_k^+,\mathbf{u}_k,\mathbf{0})} = \begin{bmatrix} 1 & 0 & -T_s v_c \sin\hat{\theta}_k^+ \\ 0 & 1 & T_s v_c \cos\hat{\theta}_k^+ \\ 0 & 0 & 1 \end{bmatrix} \quad (5)$$

$\mathbf{W}$ is the Jacobian matrix of partial derivates of $f$ to $\mathbf{w}$:

$$\mathbf{W}_k = \frac{\partial f_k}{\partial \mathbf{w}}\bigg|_{(\hat{\mathbf{x}}_k^+,\mathbf{u}_k,\mathbf{0})} = \begin{bmatrix} T_s \cos\hat{\theta}_k^+ & 0 \\ T_s \sin\hat{\theta}_k^+ & 0 \\ 0 & T_s \end{bmatrix} \quad (6)$$

$\mathbf{Q}_k$ is the input-noise covariance matrix. In the system, the input noise is modeled as being proportional to the angular speed $\omega_{L,k}$ and $\omega_{R,k}$ of the left and right wheels, respectively. Thus, the variances equal to $\delta\omega_L^2$ and $\delta\omega_R^2$, where $\delta$ is a constant with the value 0.01 determined by experiments. The input-noise covariance matrix $\mathbf{Q}_k$ is defined as:

$$\mathbf{Q}_k = \begin{bmatrix} \delta\omega_{R,k}^2 & 0 \\ 0 & \delta\omega_{L,k}^2 \end{bmatrix} \quad (7)$$

*B. The correction step with measurement update equations:*

The correction step adjusts the *priori* estimate by incorporating measurements into the estimation. The measurements in this paper are represented in form of line parameters extracted from the LRF data and deflective angular determined from the omni-directional camera.

At the first scan of the LRF, a global map of the environment, which consists of a set of line segments described by parameters $\beta_j$ and $\rho_j$, is constructed. The line equation in normal form is:

$$x_G \cos\beta_j + y_G \sin\beta_j = \rho_j \quad (8)$$

When the robot moves, a new scan of LRF is performed and a new map of the environment, namely local map, is constructed which also consists of a set of line segments described by the equation:

$$x_R \cos\psi_i + y_R \sin\psi_i = r_i \quad (9)$$

where $\psi_i$ and $r_i$ are the parameters of lines. The line segments of the global map are then transformed into the local coordinate $(X_R, Y_R)$ and match with line segments of the local map (The line segments detection and map matching algorithms will be presented in next sections). The matching line parameters $\psi_i$ and $r_i$ from the current local map are collected in the measurement vector $\mathbf{z}_k$. In order to enhance the accuracy, the robot orientation detected by the omni-camera is also added to $\mathbf{z}_k$ resulting in:

$$\mathbf{z}_k = [r_1, \psi_1, ....., r_N, \psi_N, \varphi_k]^T \quad (10)$$

This measurement vector is used as the input for the correction step of the EKF to update the robot's state (eq. 15).

The parameters $\beta_j$ and $\rho_j$ of the matched line segment from the global map (according to the global coordinates) are transformed into the parameters $\hat{\psi}_i$ and $\hat{\rho}_i$ (according to the coordinates of the robot) by

$$\hat{\rho}_i = \frac{|-k|}{\sqrt{m^2+1}} \quad (11)$$

$$\hat{\psi}_i = \arctan2(\frac{k}{1+m^2}, \frac{-mk}{1+m^2}) \quad (12)$$

where $m$ and $k$ are the slope and intercept of the line in the local coordinate as shown in (19), (20). In the present of robot's orientation, the measurement function $h$ is described by:

$$\begin{bmatrix} \hat{\rho}_i \\ \hat{\psi}_i \\ \hat{\theta} \end{bmatrix} = h(\hat{\mathbf{x}}_k^-, \beta_i, \rho_i) = \begin{bmatrix} \dfrac{|-k|}{\sqrt{m^2+1}} \\ arctan2(\dfrac{k}{1+m^2}, \dfrac{-mk}{1+m^2}) \\ \hat{\theta}^-_k \end{bmatrix} \quad (13)$$

Equations for the correction step are now described as follows:

$$\mathbf{K}_k = \mathbf{P}_k^- \mathbf{H}_k^T (\mathbf{H}_k \mathbf{P}_k^- \mathbf{H}_k^T + \mathbf{R}_k)^{-1} \quad (14)$$

$$\hat{\mathbf{x}}_k = \hat{\mathbf{x}}_k^- + \mathbf{K}_k \left( \mathbf{z}_k - h(\hat{\mathbf{x}}_k^-) \right) \quad (15)$$

$$\mathbf{P}_k = (\mathbf{I} - \mathbf{K}_k \mathbf{H}_k) \mathbf{P}_k^- \quad (16)$$

where $\hat{\mathbf{x}}_k \in \Re^n$ is the *posteriori* state estimate at step $k$ given measurement $\mathbf{z}_k$, $\mathbf{K}_k$ is the Kalman gain, $\mathbf{H}$ is the Jacobian matrix of partial derivates of $h$ to $\mathbf{x}$, $\mathbf{R}_k$ is the covariance matrix of measurement noises. In the next section, algorithms to identify parameters for the correction step are presented.

III. IDENTIFICATION OF KALMAN FILTER'S PARAMETERS FROM THE LRF AND OMNI-DIRECTIONAL CAMERA

*A. Identification of Kalman filter's parameters from the LRF*

*1) Line segments extraction*

This section describes the algorithm to extract line segments from a set of points based on the least square quadratic (LSQ) method. The point cloud was obtained from a SICK – LMS 221 laser scanner with an 180° field of view and a 0.5° angular resolution.

The data collected by the LRF was first converted into Cartesian coordination and stored in an array according to the equation:

$$x = r\cos(\varphi \dfrac{\pi}{180}) \quad (17)$$

$$y = r\sin(\varphi \dfrac{\pi}{180}) \quad (18)$$

where $r$ and $\varphi$ are the range and bearing from robot position to obstacles respectively. Points that lie too close to robot position (< *40cm*) were treated as noise and were eliminated. Because the LRF scans from right to left, neighboring points have high chance to lie on a same landmark such as on the wall. The LSQ was applied to a group of $N_1$ consecutive laser readings starting from the first points to find the best fit line of these points. A distance threshold *disT* was predefined and compared with the distance from each point in the group to the best fit line. If there were more than $N_2$ of the points in the group (in our research, $N_2 = 0.75 N_1$), the distance of which is fewer than the *disT*, a line was detected and a new best fit line was calculated based on these points. The point with the smallest index would be saved as one end of the line segment.

The slope $m$ and intercept $k$ of the line were computed using the formula:

$$m = \dfrac{\sum xy - \dfrac{(\sum x)(\sum y)}{n}}{\sum x^2 - \dfrac{(\sum x)^2}{n}} \quad (19)$$

$$k = \bar{y} - m\bar{x} \quad (20)$$

where $n$ is the total number of data points, $\bar{x}$ and $\bar{y}$ are the mean of the *x*- and *y*- coordinates of the data points respectively. The algorithm would then be used to find the remaining points of the cloud to add more points to the line segment. Points that lie too far from their previous neighbors in the array would be ignored. The last point that met the distance threshold was then treated as the other end of the found line segment. The algorithm was repeated with the new group until all elements of the collected point cloud were checked.

Nevertheless, a fixed threshold *disT* did not work for all scans. Experiment results show that in about 10% of the scans, the threshold of the distance from one point to the best fit line should be larger in order to detect lines from data points. However if the threshold applied for all scans was too large, extracted line segments would be not smooth and could affect the matching result. In this paper, a simple dynamic threshold was developed to solve this problem. A maximal threshold *disTmax* was defined as the largest distance to determine if a point belongs to a line. For each group of points, instead of comparing the distance from each point to the best fit line with fixed *disT*, all distances would be stored in an array and sorted in an ascending order. If the $N_2^{th}$ distance was smaller than the maximal threshold *disTmax*, it would be chosen as the distance threshold for this line segment. Therefore, the number of distance threshold in each scan would be the same as the number of detected line segments.

*2) Local and global maps matching*

The extracted line segments of same landmarks from local and global maps are then matched together using a straight forward algorithm. The extracted line segments from global map $G_j$ and local map $L_i$ are described as follows:

$$G_j = [x_{1G,j} \ x_{2G,j} \ y_{1G,j} \ x_{2G,j}] \quad j = 1...n_G$$
$$L_i = [x_{1L,i} \ x_{2L,i} \ y_{1L,i} \ y_{2L,i}] \quad i = 1...n_L$$

where $(x_1, y_1)$ and $(x_2, y_2)$ are Cartesian coordinate of two end points of extracted line segments respectively, $n_G$ and $n_L$ are number of extracted global and local line segments. The detected global line segments are first transformed into local map with the equation:

$$G_{Tj} = \mathbf{G}\left( \begin{bmatrix} x_{1G,j} & y_{1G,j} \\ x_{2G,j} & y_{2G,j} \end{bmatrix} - \begin{bmatrix} x_0 & y_0 \\ x_0 & y_0 \end{bmatrix} \right) \quad (21)$$

where $\mathbf{G} = \begin{pmatrix} \cos\varphi_0 & -\sin\varphi_0 \\ \sin\varphi_0 & \cos\varphi_0 \end{pmatrix}$ is the conversion matrix, $[x_o \ y_o \ \phi_o]^T$ are indicators of robot's position in global map estimated by odometry. Each local line segment $L_i$ is compared with all transformed global line segments $G_{T,j}$, and two line segments are considered as "matched" if their range and bearing to robot position are approximately the same and the overlapping rate between them is less than previous defined threshold [2].

The range and bearing of global and local line segments to robot position in robot's coordinate are presented as $(\rho_{G,j}, \psi_{G,j})$ and $(\rho_{L,i}, \psi_{L,i})$ respectively

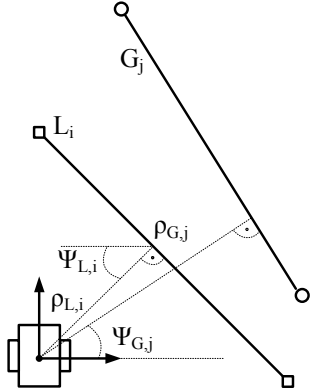

Figure 2. The range and bearing of global and local line segments to robot's position in robot's coordinate

The overlapping rate between the local line segment $L_i$ and the transformed global line segment $G_{T,j}$ is defined as follows:

$$O_k(a_k, b_k) = |a_k + b_k - \overline{G}_{T,j}|$$
$$O_k(c_k, d_k) = |c_k + d_k - \overline{G}_{T,j}|$$
$$k = (j-1)n_L + i$$
$$k = 1, 2, ..., n_L n_G - 1, n_L n_G$$

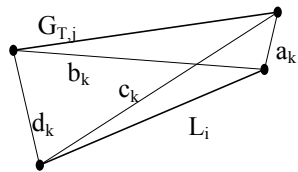

Figure 3. The overlapping parameters between global and local line segments

where $\overline{G}_{T,j}$ is the length of transformed global line segment, $a_k, b_k, c_k$ and $d_k$ are Euclidean distances between the end points of the line segments $L_i$ and $G_{T,j}$.

The inequations below represent conditions for matching local and transformed global line segments in robot's coordinate. Two line segments $L_i$ and $G_{T,j}$ are matched if all following conditions are met:

$$\begin{cases} O_k(a_k, b_k) < T \\ O_k(c_k, d_k) < T \\ (\rho_{L,i} - \rho_{G,j})^2 < T_\rho \\ (\psi_{L,i} - \psi_{G,j})^2 < T_\psi \end{cases}$$

where $T, T_\rho, T_\psi$ are predefined threshold.

*3) Estimation of line parameter's covariances*

In order to compute the measurement covariance for the EKF, each extracted line segment could be presented as $(\rho, \psi)$ parameters where $\rho$ stands for the perpendicular distance from robot's position to the line and $\psi$ is the line orientation:

$$\rho = \frac{|-k|}{\sqrt{m^2 + 1}} \quad (22)$$

$$\psi = \arctan 2\left(\frac{k}{1+m^2}, \frac{-mk}{1+m^2}\right) \quad (23)$$

where *(m, k)* are slope and intercept of detected line which were computed in (19), (20).

From the solution give by Derich [6], the measurement covariance matrix, *R*, can be calculated with the assumption that each data point has the same Cartesian uncertainty:

$$R_i = \frac{a_i \sigma_{yy}^2 - b_i \sigma_{xy}^2 + c_i \sigma_{xx}^2}{(a_i - c_i)^2 + b_i^2} \begin{pmatrix} 1 & -e \\ -e & e^2 \end{pmatrix}$$

$$+ \begin{pmatrix} 0 & 0 \\ 0 & \dfrac{\sigma_{yy}^2 \cos^2\varphi_i + \sigma_{xx}^2 \sin^2\varphi_i - 2\sigma_{xy}^2 \sin\varphi_i \cos\varphi_i}{n} \end{pmatrix} \quad (24)$$

$$\cong \begin{bmatrix} var(r_i) & 0 \\ 0 & var(\psi_i) \end{bmatrix}$$

where

$$e_i = \overline{y}\cos\psi_i - \overline{x}\sin\psi_i; \ \varphi_i = (\psi_i + \frac{\pi}{x}); \ \overline{x} = \frac{1}{n}\sum x_j; \ \overline{y} = \frac{1}{n}\sum y_j;$$

$$a_i = \sum(x_j - \overline{x})^2; \ b_i = 2\sum(x_j - \overline{x})(y_j - \overline{y}); \ c_i = \sum(y_j - \overline{y})^2.$$

In the presence of the covariance of the orientation determined by the omni-directional camera (shown in next section), the measurement noise covariance matrix for the correction step of the EKF is written as:

$$R_k \cong \begin{bmatrix} var(r_i) & ... & 0 & 0 & 0 \\ ... & var(\psi_i) & & & ... \\ ... & & ... & & ... \\ ... & & & var(\psi_i) & 0 \\ 0 & 0 & 0 & ... & 0.0265 \end{bmatrix} \quad (25)$$

*B. Identification of Kalman filter's parameters from the omni-directional camera*

In this paper, the omni-directional camera is used as an absolute orientation measurement and is fused with the LRF sensor to enhance the localization accuracy for the robot. The method is based on the detection of a red vertical landmarks located at a fixed position $(x_m, y_m)$ and paralleled with the optical axis of the camera. The conservation of line feature ensures that the shape of the landmark is unchanged in both omni-directional and panorama images. The robot's orientation determination then becomes the problem of calculating the orientation $\varphi$ of the landmark (fig.4). The algorithm can be summarized as follows: From the capture image, a digital filter is applied to eliminate random noises. The red area is then detected and the image is transformed from the RGB color space to the grey scale. By applying the Hough transform, the vertical line is extracted and the value $\varphi$ corresponding to the robot's orientation is obtained.

In order to determine the variance of the estimation, the orientation values calculated from the omni-directional camera are experimentally compared with the values measured by a high accurate compass sensor. In our system, the variance is determined with the value 0.0265 $rad^2$.

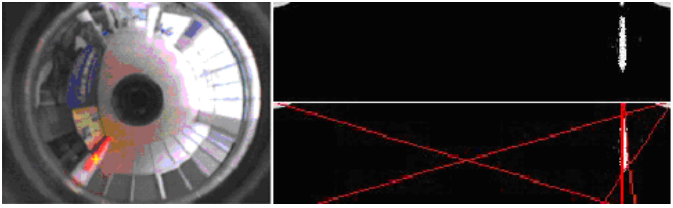

Figure 4. Line detection using Hough transform

## IV. EXPERIMENTS

In this section, experiments are conducted to evaluate the efficiency of the fusion algorithm.

### A. Experimental Setup

A rectangular shaped flat-wall environment constructed from several wooden plates surrounded by a cement wall is set up. The robot is a unicycle-like mobile robot with two wheels, differential drive (fig.5). Its wheel diameter is *10 cm* and the distance between two drive wheels is *60cm*. The sampling period $T_s$ of the EKF is *100ms*.

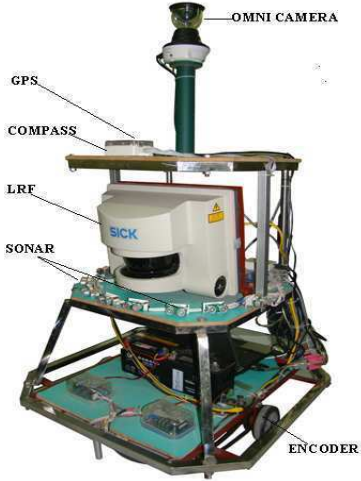

Figure 5. A unicycle-like mobile robot

### B. Line segments extraction

The algorithm was programmed in MATLAB and the computation time for a data set of 360 points with fixed threshold *disT* is 0.15s. Computation time for scanning with dynamic threshold is from 0.20s to 0.25s depending on the number of detected line segments. Fig.6 shows the result with fixed threshold *disT = 200*; fig.7 shows the result with different thresholds for different line segments.

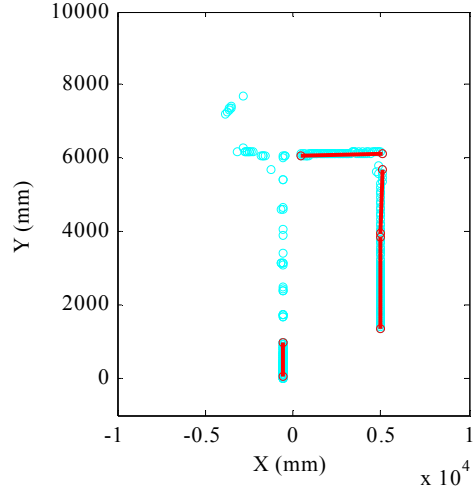

Figure 6. Extracted line segments with dynamic threshold

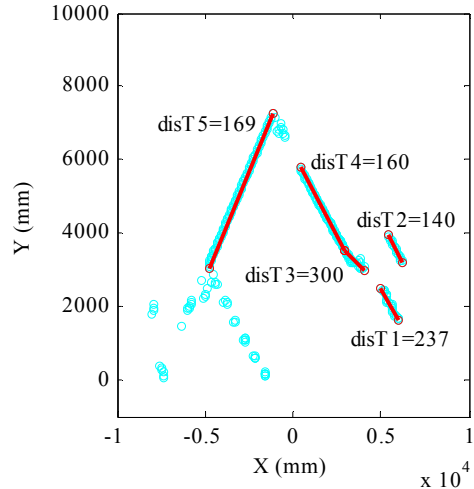

Figure 7. Extracted line segments with dynamic threshold

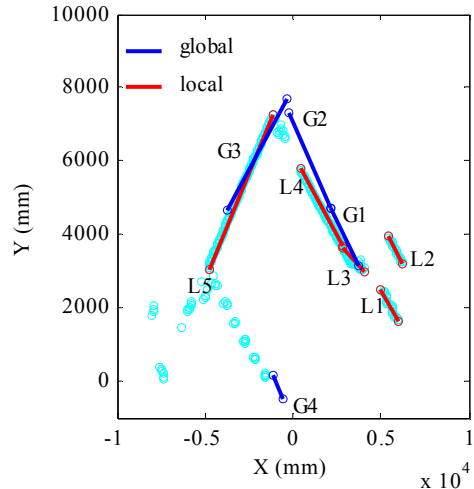

Figure 8. Matching local and global line segments

## C. Matching local and global maps

Fig.8 presents the result of the matching line segments where the blue ones stand for the transformed global line segments and the red ones are for local line segments. In reality, there are line segments detected in global map but do not appear in the local map and vice versa. Therefore, in this case, the authors only studied those appearing in both maps. With the matching criteria considered, (L3, G1), (L4, G2), (L5, G3) met all the conditions and were paired up together.

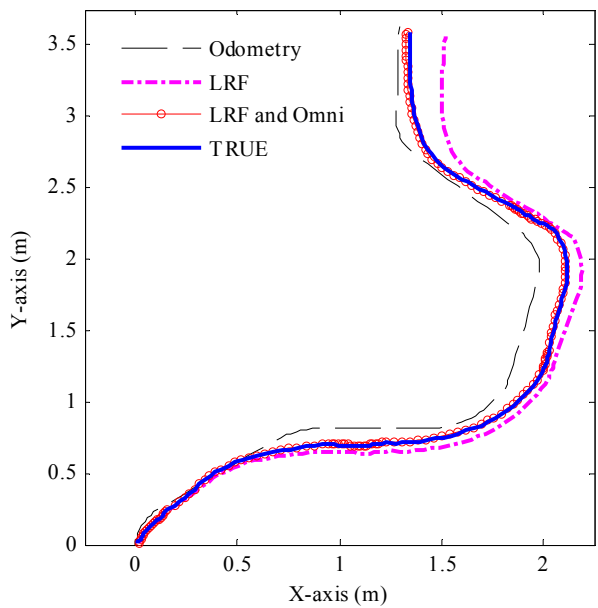

Figure 9. Estimated robot trajectories under different EKF configurations

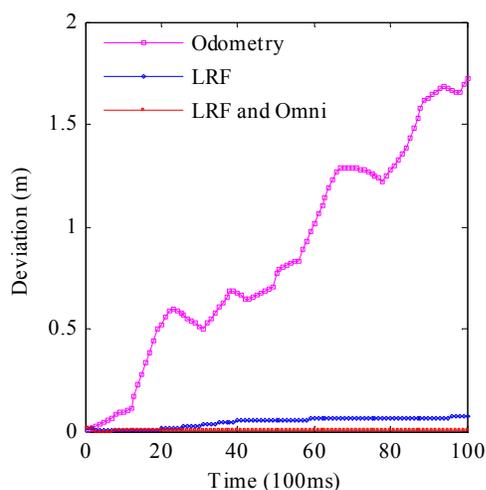

Figure 10. The deviation between estimated positions and the real one

## D. Localization using the EKF with the LRF and Omni-directional camera

In order to evaluate the efficiency of the fusion algorithm for the problem of localization, different configurations of the EKF were implemented. Fig.9 describes the trajectories of a robot movement estimated by three methods: the odometry, the the EKF with LRF, and the EKF with the combination of LRF and omni-directional camera. The deviations between each trajectories and the real one are represented in fig.10.

It is concluded that the EKF algorithm improves the robot localization and its combination with the LRF and omni-directional camera gives the *best* result.

## V. CONCLUSION

In this paper, the EKF is implemented for the problem of mobile robot localization. A combination of the LRF and omni-directional camera is introduced as the measurement system. This measurement system in combination with novel line extraction and map matching algorithms simplifies the implementation while enhancing the accuracy of the EKF. Experiments confirmed the effectiveness and applicability of the proposed approach.

## ACKNOWLEDGMENT

This work was supported by Vietnam National Foundation for Science and Technology Development (NAFOSTED).